\documentclass[journal,transmag]{IEEEtran}

\usepackage[utf8]{inputenc}
\usepackage{todonotes}
\usepackage{amsfonts}
\usepackage{tabularx}
\usepackage{array, booktabs, makecell}
\usepackage{siunitx}
\usepackage{xspace}
\usepackage{tablefootnote}
\usepackage{lipsum}
\usepackage{amsmath,amssymb}
\usepackage{hyperref}
\usepackage[font=small,labelfont=bf,tableposition=top]{caption}
\usepackage{multirow}
\usepackage{xcolor}
\usepackage[T1]{fontenc}
\usepackage{fancyhdr}

\newcommand{\Fbox}[1]{\setlength{\fboxrule}{0pt}\setlength{\fboxsep}{0pt}\fbox{#1}}

\newcommand{\simname}{{\small SUMMIT}\xspace}

\newcommand{\modelname}{{\small GAMMA}\xspace}

\newcommand{\secref}[1]{Section~\ref{#1}}
\renewcommand{\eqref}[1]{Eqn.~(\ref{#1})}
\newcommand{\figref}[1]{Fig.~\ref{#1}}

\newcommand{\tabref}[1]{Table~\ref{#1}}
\newcommand{\subfig}[1]{\textit{#1}}

\newcommand{\ie}{\textit{i.e.}}

\newcommand{\etc}{\textit{etc.}}

\newcommand{\internalstates}{human behavioral states\xspace}

\newcommand{\vo}{\ensuremath{\mathrm{VO}\xspace}} % velocity obstacle
\newcommand{\vel}{\ensuremath{v\xspace}} % velocity
\newcommand{\velu}{\ensuremath{u\xspace}} % velocity u
\newcommand{\veln}{\ensuremath{n\xspace}} % velocity n
\newcommand{\pose}{\ensuremath{\mathrm{\mathbf{p\xspace}}}} % pose
\newcommand{\timewin}{\ensuremath{\tau\xspace}} % time window
\newcommand{\responsibility}{\ensuremath{\gamma\xspace}} % responsibility

\newcommand{\attfun}{\ensuremath{\mathrm{Att}\xspace}}
\newcommand{\kinfun}{\ensuremath{f\xspace}}
\newcommand{\kinset}{\ensuremath{\mathrm{K}\xspace}}
\newcommand{\approxkinset}{\ensuremath{\hat{\mathrm{K}}\xspace}}
\newcommand{\geoset}{\ensuremath{\mathrm{G}\xspace}}
\newcommand{\conset}{\ensuremath{\mathrm{C}\xspace}}
\newcommand{\maxbounderror}{\ensuremath{\mathrm{\varepsilon_{\mathrm{max}}}\xspace}}

\newcommand{\norm}[1]{\left\lVert#1\right\rVert}
\DeclareMathOperator*{\argmin}{arg\,min} % Jan Hlavacek
\DeclareSIUnit\Ms{m/s}

\newcommand{\rvs}[1]{{#1}}

\usepackage[numbers,sort&compress]{natbib}

\def\IEEEtitletopspace{18pt}

\IEEEoverridecommandlockouts                              % This command is only needed if 
\renewcommand\IEEEkeywordsname{Keywords}

\include{setting}

\begin{document}

\title{GAMMA: A General Agent Motion Model for Autonomous Driving}

\author{Yuanfu Luo\IEEEauthorrefmark{1},
Panpan Cai\IEEEauthorrefmark{1}\IEEEauthorrefmark{2},
Yiyuan Lee, and 
David Hsu,~\IEEEmembership{Fellow,~IEEE}\\
% \thanks{Manuscript received: September, 09, 2021; Revised December, 07, 2021; Accepted January, 10, 2022.}%Use only for final RAL version
% \thanks{This paper was recommended for publication by Editor Cesar Cadena upon evaluation of the Associate Editor and Reviewers' comments.}
\thanks{
This work is supported by the National University of Singapore (AcRF grant R-252-000-A87-114).} %Use only for final RAL version
\thanks{\IEEEauthorrefmark{1}The authors contributed equally.}
\thanks{\IEEEauthorrefmark{2}Corresponding author}
\thanks{The authors are with School of Computing, National University of Singapore, Singapore 119077 (e-mail: yuanfu@comp.nus.edu.sg; pcai2@e.ntu.edu.sg; yiyuan.lee@rice.edu; dyhsu@comp.nus.edu.sg).}
% \thanks{Digital Object Identifier (DOI): see top of this page.}
}

% The paper headers
\markboth{IEEE Robotics and Automation Letters. Preprint Version. Accepted Jan, 2022}
{Luo \MakeLowercase{\textit{et al.}}: GAMMA} 

\IEEEtitleabstractindextext{%
\begin{abstract}
This paper presents \modelname, a general  motion prediction model  that enables large-scale real-time simulation  and planning  for autonomous driving.
\modelname models \textit{heterogeneous}, \textit{interactive} traffic agents that  operate under diverse road conditions, with various geometric and
kinematic constraints.
\modelname treats the prediction task as constrained optimization in traffic agents' velocity space. The objective is to optimize an agent's driving performance, while obeying  all  the constraints resulting from  the  agent’s  kinematics,  collision  avoidance  with  other  agents,  and  the  environmental  context.
Further, \modelname explicitly conditions  the  prediction on  human  behavioral  states as  parameters  of  the  optimization model,  in  order  to  account for versatile human  behaviors.
We  evaluated \modelname on  a  set  of  real-world  benchmark datasets. 
The results show that \modelname achieves high  prediction accuracy on both homogeneous and heterogeneous traffic datasets, with sub-millisecond execution time. 
Further,  the computational efficiency and the flexibility of \modelname enable (i)  simulation of  mixed urban traffic at many locations worldwide and (ii)  planning for autonomous driving in dense traffic with uncertain driver behaviors, both in real-time. The open-source code of GAMMA is available \href{https://github.com/AdaCompNUS/GAMMA}{online}. 
\end{abstract}

}

\maketitle
\IEEEdisplaynontitleabstractindextext
\IEEEpeerreviewmaketitle

% \maketitle
% \IEEEdisplaynontitleabstractindextext
% \IEEEpeerreviewmaketitle

\section{Introduction}
\begin{figure}
    \centering
    \includegraphics[width=0.45\textwidth]{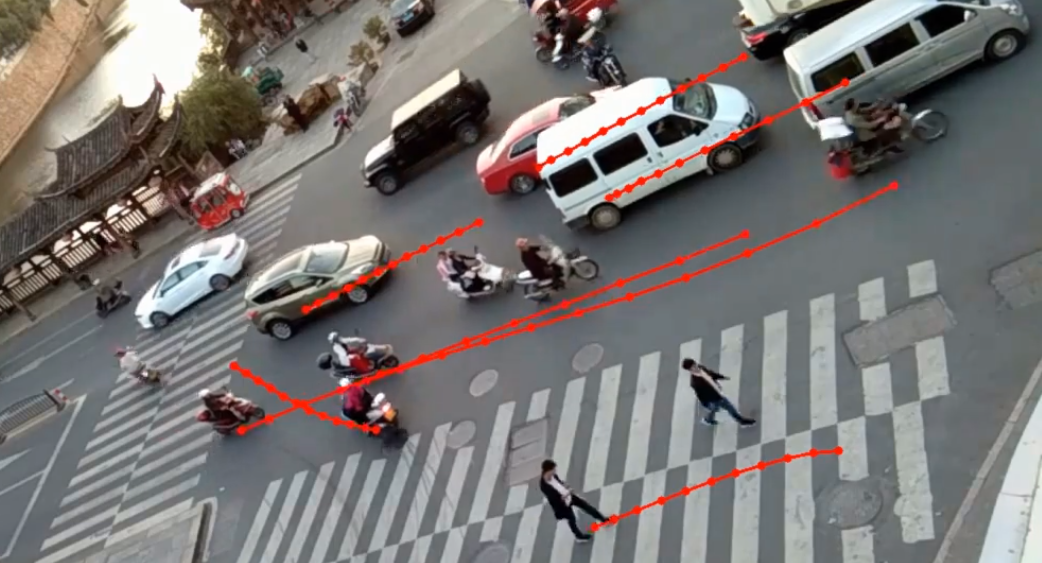}
    \caption{Motion prediction in  urban traffic.}
    \vspace{-10pt}
    \label{fig:problem}
\end{figure}

Autonomous driving holds the promise of improved safety and convenience in our daily life. However, driving through dense urban traffic, especially, on uncontrolled roads or at unsignalized intersections (\figref{fig:problem}), remains an open challenge. A core requirement of autonomous driving systems is to predict the motions of other traffic agents---pedestrians, bicycles, cars, buses, \etc---and plan for the ego-vehicle's motion accordingly.
Motion prediction, however, is very difficult, especially, with many traffic agents interacting dynamically in diverse environments. 
Traffic agents are \emph{heterogeneous}, having different geometry, kinematics, and dynamics. Human traffic participants also have different internal \textit{behavioral states}. They intend for different destinations and routes; they may be  attentive or distracted; they may act defensively or aggressively. Further, the agents react to and influence one another's behaviors.  Environmental conditions compound the complexity of agent behaviors: multi-lane roads, intersections, roundabouts, \etc. 
\rvs{Despite the challenges, \textit{accurate} and computationally \textit{efficient} prediction models are  key to
real-time traffic simulation and planning for autonomous driving.}

There are two main approaches to traffic motion prediction.  Analytical methods predict the motion by following a set of rules \cite{helbing1995social,alahi2016social,amirian2019social} or optimizing an objective \cite{ORCA,PORCA}. Data-driven methods learn prediction models from  trajectory datasets \cite{alahi2016social,gupta2018social,chandra2018traphic,yu2020spatio}. 
Analytical models require extensive domain knowledge;
data-driven models need enormous data to generalize and perform well in diverse driving scenarios. \rvs{To achieve high prediction accuracy, one may increase the model complexity or the amount of data for training, but this often degrades computational efficiency and limits   the applicability to downstream tasks, such as real-time simulation and planning.}

We propose a \textit{General Agent Motion Model for Autonomous driving} (\modelname).  \modelname  offers high prediction accuracy, computational efficiency, and simplicity of implementation. Being an analytical model, \modelname requires no training data and  handles a variety of heterogeneous, interactive traffic  in diverse urban traffic environments. 

\modelname models the motion of each traffic agent as constrained optimization in its velocity space. The objective is to maximize the agent's  driving performance: drive along an intended lane with desirable speed, while obeying all the constraints resulting from the agent's kinematics, collision avoidance with other agents, and the environmental context. By assuming polygonal agent geometry and reciprocally-optimal collision avoidance, we model the optimization problem with a quadratic objective and linear constraints, and solve it efficiently in time linear in the number of constraints. 

Furthermore, \modelname explicitly conditions the prediction on
\internalstates, such as the intended route and attention on other agents, as
parameters of the optimization model, in order to synthesize flexible human
driving behaviors.  By combining  the parameterized optimization model with
Bayesian inference, \modelname  infers \internalstates{} from the observed driving
trajectories, models their uncertainties with posterior distributions, and
makes distributional predictions. These capabilities are crucial for both
accurate motion predictions and down-stream tasks, such as  planning and
control for autonomous driving.

 We evaluate  \modelname on a set of real-world benchmark datasets for prediction
 accuracy.  In experiments, \modelname  outperforms several state-of-the-art analytical and data-driven models on both homogeneous and
 heterogeneous traffic. We further apply \modelname to two down-stream tasks. First, we 
 integrate \modelname with a high-fidelity crowd-driving simulator, \simname~\cite{cai2020summit},  to simulate the motions of many traffic agents on any urban map available in OpenStreetMap~\cite{OSM}. Second, using the simulator as a testing bed, we  develop a crowd-driving
 algorithm that leverages \modelname to model the uncertainty in motion
 prediction and plan for a robot vehicle under the uncertainty. In both cases, the
 computational efficiency of \modelname provides major benefits. The open-source codes of 
 \modelname and \simname  are  available online\footnote{\href{https://github.com/AdaCompNUS/summit}{https://github.com/AdaCompNUS/summit}\\
\hspace*{11pt}\href{https://github.com/AdaCompNUS/GAMMA}{https://github.com/AdaCompNUS/GAMMA}}.

For computational efficiency, \modelname makes several simplifying assumptions, which lead to its main limitations. 
We  assume that  \internalstates{} are static and do not change within the observation and prediction horizon. Further, they  are  independent over the traffic  agents. These assumptions  simplify the  inference of behavioral states and help \modelname to achieve real-time performance in simulation and planning, but at the cost of prediction accuracy. In applications where efficiency is not a major concern, we may relax the assumptions and combine \modelname  with more powerful Bayesian inference methods to handle dynamic changes in   behavioral states.

\section{Related Work}
\begin{figure*}[t]
%\hspace*{0cm}  
\centering
\begin{tabular}{ccc}
\includegraphics[height=1.7in]{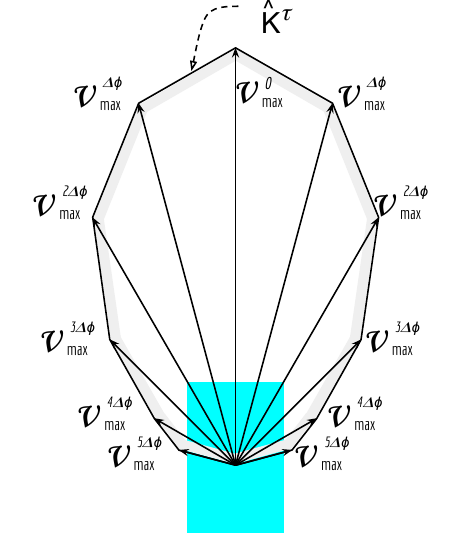}&
\includegraphics[height=1.7in]{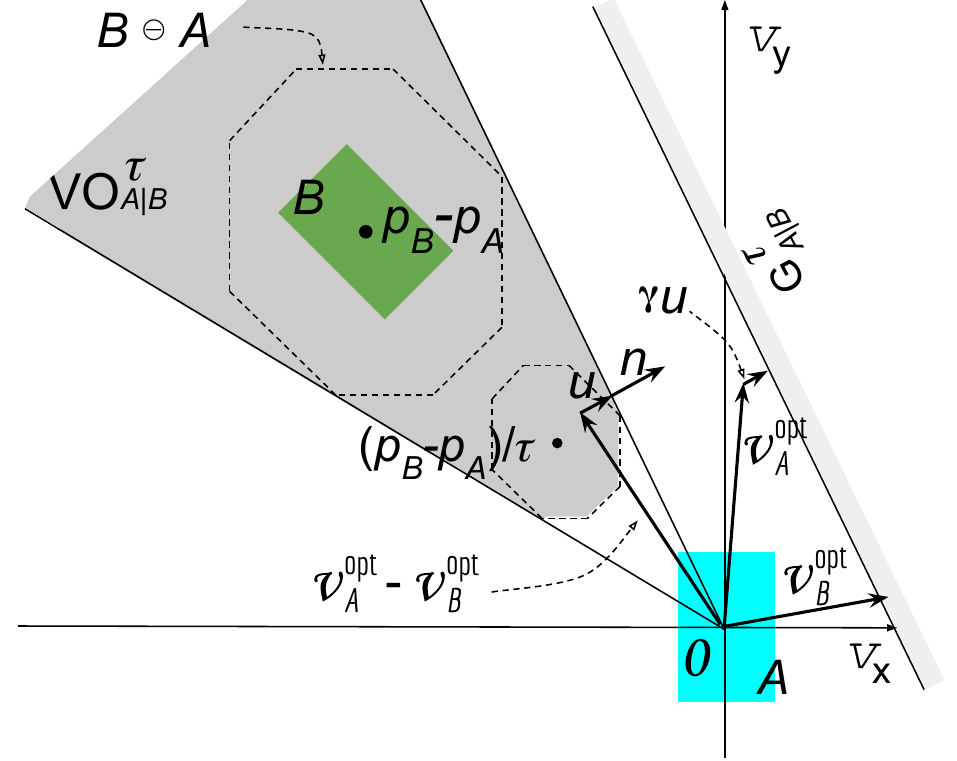} &
\hspace*{0.3cm}
\includegraphics[height=1.7in]{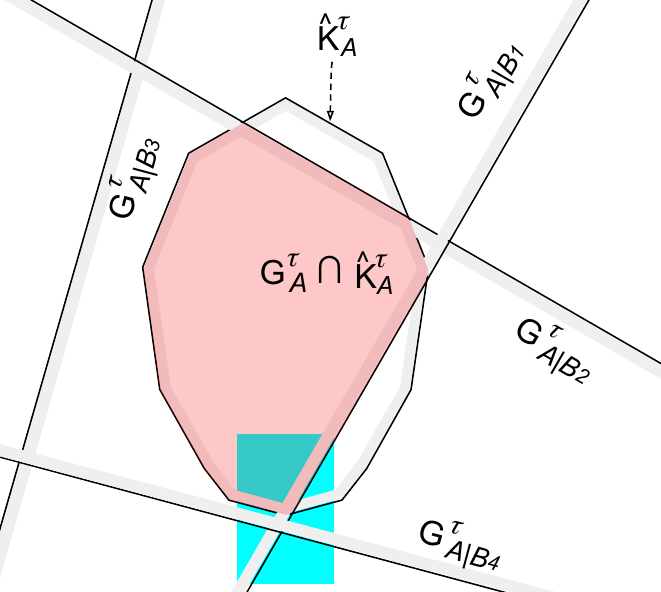} \\
      (\subfig a) & (\subfig b) & (\subfig c)

\end{tabular}
\caption{Constrained optimization over the velocity space. (\subfig a) The kinematically trackable velocity set, $\hat{\kinset}^\timewin$, estimated using the convex hull of $\{ v^\phi_{\mathrm{max}} \mid \phi \in \Phi\}$. (\subfig b) The velocity obstacle, $\vo_{A|B}^\timewin$ (gray), and the collision avoidance velocity set, $\geoset_{A|B}^\timewin$ (half plane), of agent $A$ (blue) induced by agent $B$ (green) for a given time window $\timewin$. (\subfig c) The feasible velocity set (red) of agent $A$ as the intersection of $\hat{\kinset}^\timewin_A$ and the geometry constraints induced by four other agents: $\geoset_{A|B_1}^\timewin$,$\geoset_{A|B_2}^\timewin$,$\geoset_{A|B_3}^\timewin$, and $\geoset_{A|B_4}^\timewin$. }
 \label{fig:vo-g-k-f}
     \vspace*{-10pt}
   \end{figure*}

There is a vast literature on  motion prediction. We selectively review prediction models  closely related to \modelname. 
A comprehensive survey is available in  \cite{Rudenko2020Survey}.

\subsection{Analytical Prediction Models}
Analytical methods generate motion predictions following certain rules or optimization objectives. Social force models  \cite{helbing1995social,lohner2010modeling,yamaguchi2011you} assume traffic agents to be driven by attractive forces exerted by the destination and repulsive forces exerted by obstacles. They can simulate large crowds efficiently, but the quality of interactions are often constrained by model simplicity. 
Velocity Obstacle (VO) \cite{fiorini1998motion} and Reciprocal Velocity Obstacle (RVO) \cite{van2008reciprocal} model collision avoidance behaviors of agents by solving constrained optimization problems in the velocity space. RVO-based methods can ensure collision-free constraints and are yet flexible enough to model complex interactions among traffic agents. The optimal reciprocal collision avoidance (ORCA) version \cite{van2011reciprocal} is thus chosen as the base framework of \modelname.

Later analytical approaches start to handle the heterogeneity of traffic agents in kinematics, geometry, and behavioral states of human drivers in their models.
A few RVO variants \cite{alonso2013optimal,alonso2012reciprocal,van2011reciprocalAVO} handle non-holonomic traffic agents by explicitly conditioning on kinematics or dynamics equations specified by human experts for different types of agents. While parameters of the equations are determining factors for the prediction accuracy, it would be difficult to identify them for traffic participants in reality. \modelname provides a relaxed numerical representation to better adapt to real-world traffic.

Other variants model different agent geometries by exploring more flexible representations than discs, including ellipses \cite{best2016real}, capsules \cite{stuvel2017torso}, and medial axis transformations (CTMAT) \cite{ma2018efficient}. The computational load, however, grows quickly with the complexity of the representation. We observe that polygons are sufficient to achieve tight bounding to typical traffic agents and offer the minimum trade-off of computational complexity. \modelname therefore uses polygonal representation. 

\rvs{This work is closely related to PORCA \cite{PORCA}, which predicts the interaction of pedestrians with a robot vehicle using the ORCA model~\cite{ORCA}.
\modelname makes several new advances in several aspects. 
PORCA only predicts for pedestrians, which has homogeneous, holonomic kinematics and disc-like shapes. It only supports pre-defined goal locations as intentions, restricting it to off-road, semi-open spaces. 
\modelname supports heterogeneous agents in on-road environments with varying geometrical features and distinct kinematics. It uses a richer polygonal representation to model the agent's geometry  and also explicitly models the kinematics constraints. \modelname additionally allows both contextual and non-contextual representations of intentions and uses explicit contextual constraints to condition predictions on arbitrary road structures. 
}

\subsection{Data-Driven Prediction Models}
A parallel line of research is to learn prediction models from real-world traffic data, typically using LSTMS or graph neural networks (GNNs). LSTM-based models usually include an encoder-decoder structure with an interaction module operating in the latent feature space.
Existing interaction modules include pooling operations \cite{alahi2016social,gupta2018social}, message passing \cite{zhang2019sr}, attention modules \cite{sadeghian2019sophie,amirian2019social}, and spatial convolutions \cite{chandra2018traphic,zhao2019multi}. GNN models explicitly model the interaction among agents as spatial-temporal graphs, and apply graph learning techniques to capture interactions, e.g., self-attentions \cite{yu2020spatio} and sparse graph learning \cite{shi2021sgcn}.
Heterogeneity of agents is handled by using heterogeneous training data \cite{zhao2019multi} and injecting heterogeneity features
as part of agent states and input to the models \cite{chandra2018traphic, ma2018trafficpredict}.  
Recent data-driven models also support distributional outputs,  by learning bi-variate Gaussian distributions over future positions \cite{alahi2016social,chandra2018traphic,ma2018trafficpredict,shi2021sgcn}, or training with generative-adversarial networks (GANs) \cite{gupta2018social,sadeghian2019sophie,zhao2019multi}.  

While data-driven models  leverage real-world experiences, the main challenges are generalization and \rvs{efficiency}: it requires an enormous amount of data to handle the diversity of urban roads, driving scenarios, and traffic patterns; \rvs{a strong model typically contains many parameters and requires heavy computation.} Instead, \modelname uses a simple and efficient analytical model and relies on online inference, instead of offline learning, to tackle this diversity.

\section{GAMMA} \label{sec::gamma}
\modelname assumes each traffic agent optimizes its velocity based on the navigation intention, while being constrained by kinematics, geometries, and road contexts.

A velocity $\vel$ for a traffic agent $A$ \rvs{(define in the world frame)} is called \emph{kinematically trackable} if $A$ can track $\vel$ with its low-level \rvs{PID} controller while keeping the tracking error below a threshold  $\maxbounderror$ for $\timewin$ time.
It is further called \emph{geometrically feasible} if it does not lead to collisions with any other nearby agents by taking $\vel$ for $\timewin$ time. 
Finally, $\vel$ is called \emph{contextually conformed} if it does not violate the hard contextual constraints for $\timewin$ time. The cooresponding constrained velocity sets are denoted as $\kinset_A^\tau$, $\geoset_A^\tau$ and $\conset_A^\tau$, respectively.
\modelname selects a new velocity for $A$, $\vel_A^\mathrm{new}$, from the intersection of the three constraints, $\kinset_{A}^{\tau} \cap \geoset_{A}^{\tau} \cap \conset_{A}^{\tau}$, which best aligns with the intention of the agent. 
This requires solving a constrained optimization problem. 

In the following, we will first present the modeling of the three constraints in Sections \ref{sec:kinematics}, \ref{sec:geometry} and \ref{sec:context}, then the optimization objective in \secref{sec:objective}, and finally, the inference of \internalstates{} in \secref{sec:bayesian-inference}.

\subsection{Kinematics Constraints} \label{sec:kinematics}

Kinematics significantly influence the behaviours of real-world traffic agents: pedestrians undertake flexible, holonomic motion; vehicles have non-holonomic kinematics thus cannot move sidewise.
Such constraints are typically represented as differential equations.
\modelname relaxes the representation by allowing all kinematically-trackable velocities under a maximum tracking error $\maxbounderror$ and a time window $\tau$.   

Concretely, for a given traffic agent $A$, we aim to find the \emph{kinematically trackable velocity set}, $\kinset^\timewin_A$, \emph{s.t.},
\begin{equation}\label{eq:kinematic}
\kinset^\timewin_A = \{ \vel \mid \norm{\vel t - \kinfun_A(\vel,t)} \leq \maxbounderror , \forall t \in [0, \timewin] \},
\end{equation}
where $\kinfun_A(\vel,t)$ is $A's$ future position at time $t$ if it tracks $\vel$ with its low-level controller. As $\kinfun_A(\vel,t)$ varies for different kinematics and controllers, it is often intractable to compute its analytic form. We propose to estimate $\kinset^\timewin_A$ numerically. 

We discretize velocities according to the speed and the deviation angle from an agent's heading direction by constructing a discrete set of values for each dimension: $\hat{S} = [0:\Delta s:s_\mathrm{max}]$ and $\hat{\Phi} = [0:\Delta \phi:\phi_\mathrm{\max}]$.
For each deviation angle $\phi \in \hat{\Phi}$, we
scan through the target speeds $s \in \hat{S}$ to estimate the maximum tracking error of running the low-level controller for $\timewin$ time, denoted as $\varepsilon_{(s,\phi)}$.
We then pick the maximum speed along $\phi$ that satisfies $\varepsilon_{(s,\phi)}<\maxbounderror$, and record it as a boundary point of $\kinset^\timewin$, denoted as $v^\phi_{\mathrm{max}}$. After scanning through all deviation angles, we approximate $\kinset^\timewin$ using the convex hull shaped by the boundary velocities (\figref{fig:vo-g-k-f}\subfig{a}):
\begin{equation}
\approxkinset^\timewin = \mathrm{ConvexHull}(\{ \vel^\phi_{\mathrm{max}} \mid \phi \in \Phi\})
\end{equation}

Construction of $\kinset^\timewin$ is conducted offline for each representative type of agent: pedestrian, bicycle, motorbike, car, van, bus, gyro-scooter, trucks, etc.
At run-time, agent types can be recognized by traffic detection modules \cite{YOLO_traffic}. We assume them to be known in our experiments.

\subsection{Geometric Constraints}\label{sec:geometry}
\modelname ensures collision avoidance using the following procedure. It first considers the interaction of an agent $A$ with every single agent $B$ in its neighborhood to construct a collision avoidance velocity set.
The final \textit{geometrically feasible velocity set} of $A$ is computed by intersecting the set for all $B's$. Our definition of constraint extends that in POCRA \cite{PORCA} by replacing disc-shaped agents with polygons, which offer tighter fits to typical traffic agents.

Consider a pair of agents $A$ and $B$ at positions $\pose_A$ and $\pose_B$, respectively. To simplify reasoning, we choose the ego-centric coordinate system at $A$ and convert $A$ to a single point by inflating the geometry of $B$ using that of $A$. Then, we can compute the \textit{velocity obstacle} as the set of \emph{relative velocities} that \textit{lead to} collisions before time $\timewin$, from the \textit{Minkowski difference} of the agent polygons, $B \ominus A$:
\begin{equation}
\label{eq:vo-general}
\vo_{A|B}^\timewin = \{\vel_{A|B} \mid \exists t \in [0,\timewin], t \cdot \vel_{A|B} \in (B \ominus A) \},
\end{equation}
where $\vel_{A|B}$ is a relative velocity of $A$ w.r.t. $B$. \figref{fig:vo-g-k-f}\subfig{b} visualizes a velocity obstacle constructed for two box-shaped agents.

\modelname assumes agents to be reciprocally optimal, seeking to collaboratively avoid the collision obstacle with minimum adaptation from the current relative velocity, $\vel^{cur}_{A|B}=\vel_A^\mathrm{cur}-\vel_B^\mathrm{cur}$. 
In the trivial case when $\vel^{cur}_{A|B}$ does not intersect with the velocity obstacle, the velocities can be safely maintained. Otherwise, \modelname finds the closest relative velocity to $\vel^{cur}_{A|B}$ from the boundary of the velocity obstacle as the target:
\begin{equation} \label{eq:velu}
\vel^*_{A|B} = \argmin_{\vel \in \partial \vo_{A|B}^\timewin} \norm{\vel-\vel^{cur}_{A|B} }.
\end{equation}
$\velu=\vel^*_{A|B}-\vel^{cur}_{A|B}$ is thus the smallest change on the relative velocity of $A$ and $B$ to avoid collision within $\timewin$ time.

\modelname lets $A$ take $\responsibility{} \in [0,1]$ of the \textit{responsibility} for collision avoidance by adapting its velocity by \textit{at least} $\responsibility{} \cdot \velu$. 
This defines the desirable velocity set as a half-plane:
\begin{equation}
\label{eq:geo_A_B}
\geoset_{A|B}^\timewin = \{ \vel_A \mid (\vel_A - (\vel_A^\mathrm{cur}+\responsibility{} \cdot \velu))\cdot \veln \geq 0 \},
\end{equation}
where $\veln$ is the direction of $\velu$, and $\responsibility$ denotes the minimum responsibility that $A$ is willing to take.
The responsibility level can be either fixed as $0.5$ or  be inferred from the agent's history (\secref{sec:bayesian-inference}). \figref{fig:vo-g-k-f}\subfig{b} visualizes $\geoset_{A|B}^\timewin$ induced by two box-shaped agents. 

Finally, the geometrically feasible velocity space of $A$ is constructed by considering all nearby agents:
\begin{equation}
\label{eq:geo_A}
\geoset{}_{A}^{\timewin} = \bigcap_{B \in \attfun{}(A)} \geoset{}_{A|B}^{\timewin}
\end{equation}
where $\attfun{}(A)$ is the attentive range of $A$.
We use two half circles to model agent's attention: one in front of the vehicle with radius $r_\mathrm{front}$, and one at the back with radius $r_\mathrm{rear}$. We set $r_\mathrm{rear} \leq r_\mathrm{front}$, such that agents give more attention to others in the front. The actual values of $r_\mathrm{front}$ and $r_\mathrm{rear}$ can also be inferred from history (\secref{sec:bayesian-inference}). 
\figref{fig:vo-g-k-f}\subfig{c} visualizes an example of $\geoset{}_{A}^{\timewin}$ induced by four other agents, overlaid on $\kinset_A^\timewin$.

\subsection{Contextual Constraints}\label{sec:context}

\modelname models road structures and traffic rules by injecting an additional set of \emph{contextual constraints}, denoted as $\conset_A$ for an agent $A$. 
Take driving direction as an example.
When an agent $A$ is at a distance $d$ from the lane boundary,  to avoid interfering with the opposite lane,  the agent should constrain its lateral speed under $d/\tau$.
This induces an additional half-plane constraint in the velocity space, $\conset_{A|Lane}^{\tau}$, defined by a separation line tangential to the lane boundary and with an offset of $d/\tau$ from the origin.

\subsection{Optimization Objective}  \label{sec:objective}
\modelname assumes an agent $A$ always picks from the feasible set, $\kinset_{A}^{\tau} \cap \geoset_{A}^{\tau} \cap \conset_{A}^{\tau}$,  an optimal velocity closest to its \textit{preferred velocity}, $\vel_A^\mathrm{pref}$:
\begin{equation}\label{eq:gamma-obj-fun}
v_A^\mathrm{new} = \argmin_{v \in \kinset_{A}^{\tau} \cap \geoset_{A}^{\tau} \cap \conset_{A}^{\tau}} \norm{v - v_A^\mathrm{pref}}
\end{equation}

We derive the preferred velocity from the agent's intention, which is defined differently under different contexts.
\rvs{When data on road structures are accessible, we define agent intentions as available paths to proceed on the lane network or the sidewalk network. A preferred velocity thus points towards a look-ahead sub-goal along a selected path, which is updated at each prediction step. Such contextual intentions naturally induce common maneuvers in daily driving, such as lane-keeping, lane change, overtaking, \etc.}
When road contexts are not available, like for the datasets used in our experiments, we adopted non-contextual intentions characterizing whether an agent wants to keep its current velocity or current acceleration. 
Given the above parameterization, we apply Bayesian inference (\secref{sec:bayesian-inference}) to infer the actual intentions of real-world agents. 

The objective function (\eqref{eq:gamma-obj-fun}) is quadratic, and the feasible set $\geoset{}_{A}^{\timewin}\cap \kinset{}^\timewin_A\cap C_{A}^{\timewin}$ is convex by construction. The optimization problem can therefore be efficiently \rvs{constructed and} solved in linear time w.r.t the number of agents involved. The actual motion of $A$ is generated by tracking $\vel_A^\mathrm{new}$ using the low-level \rvs{PID} controller of the particular agent type.
\rvs{The complexity of predicting for a traffic scene is thus $O(NM)$, where $N$ is the total number of agents in the scene and $M$ is the average number of neighbors for each agent.}

\subsection{Inference of Human Behavioral States}
\label{sec:bayesian-inference}
The above formulation induced a set of \internalstates{} on which \modelname is conditioned, including the intention of an agent $A$ that determines the preferred velocity, the attention range of $A$, $\attfun{}(A)$, and the collision avoidance responsibility, \responsibility{}, that agent $A$ is willing to take. These \internalstates{} can significantly change the behaviors of agents, but are not directly observable. We propose to infer them using online Bayesian inference. 

Particularly, given the observed trajectories of a set of agents in the scene, we aim to infer a \textit{belief}, or a probability distribution, over the \internalstates{} of all agents. We assume that the behavioral states of agents are static and mutually independent during the observed interactions. Consequently, a joint belief can be simplified as a collection of discrete distributions over behavioral states for each agent. A uniform prior is constructed for each new agent that appears. We then apply a factored histogram filter to track the belief using online observations.  

At each time step, for a particular hypothesis of behavioral state $\theta$ of a particular agent, we assume the agent's \textit{expected} next position to be:
\begin{equation}
\label{eq:dynamic-model}
\bar{\pose}_{t} = \mathrm{\modelname}(\theta, \pose_{t-1}),
\end{equation}
where $\pose_{t-1}$ is the history position at time $t-1$, and $\bar{\pose}_{t}$ is the next position predicted by \modelname conditioned on $\theta$. 

After observing the \textit{actual} next position of the agent, $\pose_{t}$, its likelihood w.r.t. hypothesis $\theta$ is computed as:
\begin{equation}
\label{eq:motion-dynamic-model}
p(\pose_{t} \mid \theta) = f(\norm{\pose_{t} - \bar{\pose}_{t}} \mid 0, \sigma^2),
\end{equation}
where we assume the transition noise to be Gaussian, specified by a probability density function $f$ with zero mean and a predefined variance $\sigma^2$. 

Using this likelihood, we update the posterior distribution over $\theta$ using the Bayes' rule:
\begin{equation}
\label{eq:bayes-rule-his}
p_{t}(\theta) = \eta \cdot p(\pose_{t} \mid \theta) \cdot p_{t-1}(\theta),
\end{equation}
where $\eta$ is the normalization constant. 

This process is repeated for all agents and all combinations of behavioral states to update the joint belief at each time step. We then use the updated belief to condition predictions, by solving the constrained optimization problem in Eqn. (\ref{eq:gamma-obj-fun}) conditioned on sampled \internalstates{}. In our experiment, we test both deterministic predictions using the maximum-likelihood state and distributional predictions by sampling from the posterior. 

\section{Experimental Results}

\begin{table*}[!t]
\centering
\caption{Performance comparison on benchmark datasets with \textit{homogeneous} traffic. Each entry reports the prediction ADE/FDE (\textit{m}).  
}
\label{tab:learning-based-approaches}
\scriptsize
\begin{tabular}{lrrrrrrrrrrrr} 
\toprule
\multicolumn{1}{c}{} & \multicolumn{6}{c}{\textbf{\textbf{Deterministic Models}}}                                                                                 & \multicolumn{6}{c}{\textbf{Distributional Models}}                                                      \\ 
Dataset
              &  \multicolumn{1}{l}{SRLSTM} & \multicolumn{1}{l}{STAR} & \multicolumn{1}{l}{SGAN-avg} & \multicolumn{1}{l}{S-Force}   & \multicolumn{1}{l}{PORCA} & \multicolumn{1}{l}{GAMMA} & \multicolumn{1}{l}{SGAN} & \multicolumn{1}{l}{SoPhie} & \multicolumn{1}{l}{MATF} & \multicolumn{1}{l}{S-Ways} & \multicolumn{1}{l}{SGCN} & \multicolumn{1}{l}{GAMMA-d}  \\
\cmidrule(lr){2-7}\cmidrule(lr){8-13}
ETH                  & 0.63/1.25                  & 0.56/1.11                & 0.96/1.87            & 0.67/1.52 & 0.52/1.09                 & \textbf{0.51/1.08}        & 0.81/1.52                & 0.70/1.43                  & 1.01/1.75                & 0.39/0.64                  & 0.63/1.03                & \textbf{0.30/0.65}           \\
HOTEL                & 0.37/0.74                  & \textbf{0.26/0.50}       & 0.61/1.31            & 0.52/1.03 & 0.29/0.60                 & 0.28/0.59                 & 0.72/1.61                & 0.76/1.67                  & 0.43/0.80                & 0.39/0.66                  & 0.32/0.55                & \textbf{0.18/0.40}           \\
UNIV                 & 0.51/1.10                  & 0.52/1.15                & 0.57/1.22           & 0.74/1.12 & 0.47/1.09                 & \textbf{0.44/1.06}        & 0.60/1.26                & 0.54/1.24                  & 0.44/0.91                & 0.55/1.31                  & 0.37/0.70                & \textbf{0.32/0.79}           \\
ZARA1                & 0.41/0.90                  & 0.41/0.90                & 0.45/0.98           & 0.40/0.60 & 0.37/0.87                 & \textbf{0.36/0.86}        & 0.34/0.69                & 0.30/0.63                  & 0.26/0.45                & 0.44/0.64                  & 0.29/0.53                & \textbf{0.24/0.57}           \\
ZARA2                & 0.32/0.70                  & 0.31/0.71                & 0.39/0.86           & 0.40/0.68 & 0.30/0.70                 & \textbf{0.28/0.68}        & 0.42/0.84                & 0.38/0.78                  & 0.26/0.57                & 0.51/0.92                  & 0.25/0.45                & \textbf{0.19/0.46}           \\
\bottomrule
\end{tabular}
\vspace{-6pt}
\end{table*}

Our experiments investigate the performance of  \modelname on real-world traffic datasets and compare it with  state-of-the-art  analytical and data-driven models (Sections \ref{sec:pedestrian-datasets} and \ref{sec:heterogeneous-agents-datasets}). We  conduct an ablation study on \modelname to identify the key contributors to its performance (\secref{sec:ablation-study}). Finally, we compare \modelname  with data-driven models on computational efficiency (\secref{sec:speed-comparison}).

\subsubsection{Datasets} We test \modelname and baselines on two \textit{homogeneous} benchmark datasets, ETH and UCY, and two \textit{heterogeneous} datasets, UTOWN and CROSS.
ETH consists of two scenes: a plaza outside the ETH building (ETH) and a street outside a hotel (HOTEL); 
UCY also consists of two scenes: a university plaza (UNIV) and a street outside Zara (ZARA1 and ZARA2). 
These two datasets contain only pedestrians. 
We further collect two new datasets, UTOWN and CROSS, on heterogeneous scenes, including a campus plaza with many pedestrians interacting with a vehicle scooter and an un-signalized cross in China containing many types of agents, including cars, vans, buses, bicycles, motorcycles, electric-tricycles, and pedestrians. We have released these datasets via \texttt{\url{https://github.com/AdaCompNUS/GAMMA}}.

\subsubsection{Baselines} We compare \modelname with the following analytical and data-driven models:
\begin{itemize}
   \item{Analytical models: \textit{Social force}, \textit{PORCA}.} Social force is taken from \cite{yamaguchi2011you}; PORCA \cite{PORCA} is an recent RVO-based method designed for pedestrian-vehicle interactions.
   \item{Data-driven models with deterministic predictions: \textit{SR-LSTM}, and \textit{STAR}.} SR-LSTM \cite{zhang2019sr} is built using LSTMs; STAR\cite{yu2020spatio} is based on GNN.
   \item{Data-driven models with distributional predictions: \textit{SGAN}, \textit{SoPhie}, \textit{MATF}, \textit{S-Ways}, \textit{SGCN}}. SGAN \cite{gupta2018social}, SoPhie \cite{sadeghian2019sophie}, MATF \cite{zhao2019multi}, and S-Ways \cite{amirian2019social} are trained using GAN. SGCN \cite{shi2021sgcn} is trained to predict Gaussian;
\end{itemize}

\subsubsection{Evaluation} \rvs{We evaluate \modelname and baseline models using the same convention as previous works \cite{alahi2016social,zhang2019sr,ma2018trafficpredict,yu2020spatio,gupta2018social,sadeghian2019sophie,zhao2019multi,amirian2019social}. For each comparison, we evaluate all relevant models using standardized metrics brought forward from the cited works, including: 
\begin{itemize}
   \item{Average Displacement Error (\textbf{ADE}).} The Euclidean distance (in meters) between the predicted and ground-truth positions averaged over the entire prediction horizon. 
   \item{Final Displacement Error (\textbf{FDE}).} The Euclidean distance (in meter) between the predicted and ground-truth positions at the final frame (the $12th$). 
  As the error accumulates over time, FDE generally reflects the worst-case error over the prediction horizon.
\end{itemize}
All models accept the same dataset format and predict for 12 frames (4.8s), each time step corresponding to $0.4s$. On ETH and UCY, we directly use the performance scores of data-driven models reported by the original papers; On UTOWN and CROSS, we evaluate the released pre-trained models of data-driven baselines, if available, using their released evaluation code and with the same metrics.
}

\subsection{Prediction Accuracy in Homogeneous Traffic}
\label{sec:pedestrian-datasets}
We first test \modelname on pedestrian datasets, ETH and UCY, on which most exiting data-driven models are trained.
To conduct a fair comparison with both deterministic and distributional models, we also consider two versions of \modelname: a deterministic version that uses the most-likely behavioral state to generate a prediction, and a distributional version (\modelname-d) that samples 20 hypotheses from the posterior distribution and chooses the best prediction. The distributional measure is consistent with the best-in-20 measure used by SGAN, SoPhie, MATF, and S-Ways. 
It is worth noticing that picking the best prediction from many does not reflect a model's advantage for application in driving. It is often the expected accuracy that matters. We therefore also include SGAN-avg, which measures the average accuracy of all predictions from SGAN, as an additional baseline.

\tabref{tab:learning-based-approaches} shows the ADE/FDE of \modelname and other models. \modelname outperforms data-driven models in both deterministic and distributional setups, without consuming any training data.
\modelname also significantly outperforms social force in complex scenes like HOTEL and UNIV.  
The improvement over PORCA is not salient here as the scenes contain only pedestrians. 
However, we will show later in \secref{sec:heterogeneous-agents-datasets} that \modelname outperforms PORCA significantly on heterogeneous datasets.

\begin{figure} [!t]
\centering
\begin{tabular}{ccc}
\hspace{-8pt}
\includegraphics[height=0.22\columnwidth]{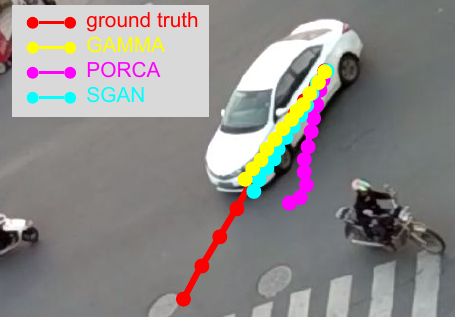} & 
\hspace{-13pt}
\includegraphics[height=0.22\columnwidth]{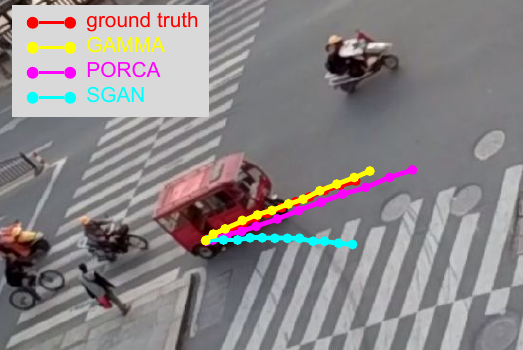} & \hspace{-13pt}
\includegraphics[height=0.22\columnwidth]{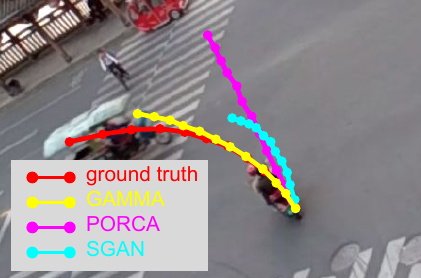} \\
\scriptsize (\subfig{a}) & \hspace{-10pt}\scriptsize (\subfig{b}) & \hspace{-15pt}\scriptsize (\subfig{c}) \\
% \vspace{-22pt}
\end{tabular}
\vspace{-3pt}
\caption{
The  predictions of various models, compared with the ground truth. (\subfig{a})~The PORCA prediction  is  infeasible. (\subfig{b})~The SGAN prediction   points to a wrong direction. (\subfig{c})~The \modelname prediction is closest to the ground truth.
}\label{fig:prediction-visualization}
\vspace{-10pt}
\end{figure}

\begin{figure} [!t]
    \centering
    \includegraphics[width=0.45\textwidth]{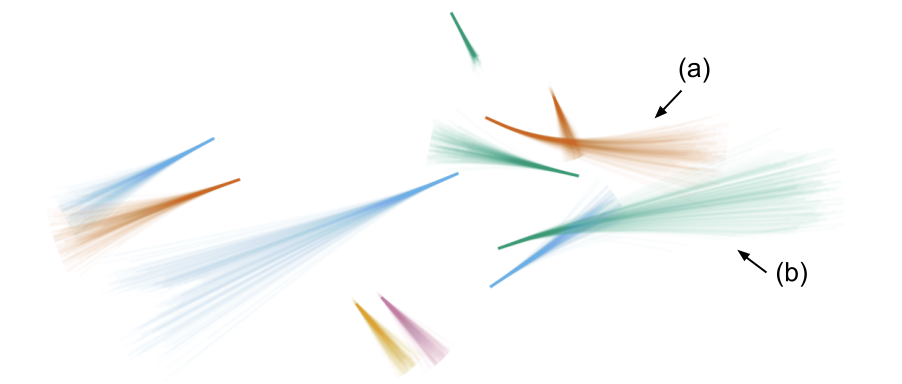}
    \caption{Distributional predictions from \modelname visualized as trajectories generated using sampled \internalstates{} from the posterior. Colors are used to assist the visuals of distinct agents (reuse is allowed). (\subfig{a}) and (\subfig{b}) show two agents interacting reciprocally and with others. The predictions are safe, smooth, and multi-modal.}
    \label{fig:distribution_vis}
    \vspace{-10pt}

\end{figure}

\subsection{Prediction Accuracy in Heterogeneous Traffic}
\label{sec:heterogeneous-agents-datasets}

\begin{table}
\centering
\caption{Performance comparison on benchmark datasets with  \textit{heterogeneous} datasets. Each entry reports the ADE/FDE/COL. \rvs{``COL'' represents the collision rate, \ie, the ratio of agent pairs that overlap with each other across all frames.}}
\label{tab:traditional-approaches}
\scriptsize
\resizebox{\columnwidth}{!}{% <------ Don't forget this %

\begin{tabular}{lrrrr} 
\toprule
Dataset & \multicolumn{1}{l}{SRLSTM}  & \multicolumn{1}{l}{SGAN} & \multicolumn{1}{l}{PORCA} & \multicolumn{1}{l}{GAMMA}    \\
\midrule
CROSS   & 1.36/3.17/\rvs{0.16}                  & 0.93/2.18/\rvs{0.30}                & 0.89/2.10/\rvs{\textbf{0.005}}                 & \textbf{0.64}/\textbf{1.77}/\rvs{0.02}  \\
UTOWN   & 0.41/0.96/\rvs{0.13}                   & 0.59/1.43/\rvs{0.15}                & 0.34/0.86/\rvs{\textbf{0.004}}                 & \textbf{0.32}/\textbf{0.84}/\rvs{0.03}  \\
\bottomrule
\end{tabular}%
}
\end{table}

We further test \modelname on heterogeneous datasets, UTOWN and CROSS. For comparison, we use PORCA as the strongest analytical model, together with data-driven models including SR-LSTM with deterministic predictions, and SGAN with distributional predictions. 

\tabref{tab:traditional-approaches} shows that \modelname significantly outperforms PORCA and data-driven models, especially for CROSS where heterogeneous traffic agents interact intensively with others. While the generalized performances of data-driven models degrade in the new environment, \modelname achieves similar performance as in homogeneous scenes. \rvs{\modelname also generates much fewer collisions than the data-driven baselines, further indicating improved realism.} 
\figref{fig:prediction-visualization} shows sample predictions from \modelname, PORCA, and SGAN contrasted to the ground-truth trajectories.
GAMMA correctly identified agents' intentions and generated smooth trajectories better conformed to vehicle geometry and kinematics.
\figref{fig:distribution_vis} visualizes distributional predictions from \modelname for a traffic scene in CROSS. \modelname provides safe and smooth trajectories, and in the meantime, captures the uncertainty over possible ways to interpret the history and to achieve collision avoidance.
More prediction results on the CROSS dataset can be found in the accompanying video (\href{https://youtu.be/teZJWlh8ZqI}{link}).

\subsection{Ablation Study}
\label{sec:ablation-study}
\begin{table}[t]
\centering
\caption{Ablation study of \modelname on the  CROSS dataset.  Each entry reports the prediction ADE/FDE. }\label{tab:ablation-study}
{\scriptsize
\begin{tabular}{cccc}
\toprule
\multicolumn{1}{c}{\begin{tabular}[c]{@{}c@{}}GAMMA \\ \mbox{}\end{tabular}}     &
\multicolumn{1}{c}{\begin{tabular}[c]{@{}c@{}}w/o kinematic \\constraints\end{tabular}}  \hspace{-5pt}& \multicolumn{1}{c}{\begin{tabular}[c]{@{}c@{}}w/o polygon\\ representation\end{tabular}}  \hspace{-5pt}& \multicolumn{1}{c}{\begin{tabular}[c]{@{}c@{}}w/o intention \\inference\end{tabular}}  \hspace{-5pt}     \\ 
\cmidrule(lr){0-3}
\textbf{0.64}/\textbf{1.77} & 0.68/1.84   \hspace{-5pt}& 0.84/2.07  \hspace{-5pt}& 0.96/2.29 \hspace{-5pt} \\
\bottomrule
\end{tabular}
}
\vspace{-5pt}
\end{table}

We further conduct an ablation study on the core innovations of \modelname by disabling the kinematics constraints, polygonal geometry, or intention inference.
Particularly, \modelname w/o kinematics assumes holonomic agents; \modelname w/o polygon representation models agents as discs; and \modelname w/o intention inference assumes agents to take a fixed, current velocity intention. \tabref{tab:ablation-study} presents the results evaluated using the CROSS dataset, showing that each of the tested components contributes significantly to the prediction accuracy.

\subsection{Computational Efficiency}
\label{sec:speed-comparison}

\begin{table}[!t]
\centering
\caption{Comparison of  execution time ( \textit{ms}) of \modelname  with data-driven models SGAN and SRLSTM with different batch sizes (bs). All results are measured on a workstation with Intel Core i7-7700K CPU and Nvidia GTX 1080 GPU. }\label{tab:speed-comparison}
{\scriptsize
\begin{tabular}{crrrrr}
\toprule
         & \begin{tabular}[c]{@{}c@{}}SGAN \\ ($bs=1$)\end{tabular} & \begin{tabular}[c]{@{}c@{}}SGAN \\ ($bs=64$)\end{tabular} & \begin{tabular}[c]{@{}c@{}}SRLSTM\\ ($bs=1$)\end{tabular} & \begin{tabular}[c]{@{}c@{}}SRLSTM\\ ($bs=4$)\end{tabular} & 
         \begin{tabular}[c]{@{}c@{}}GAMMA\\ \mbox{}\end{tabular}   \\ %\hline \hline
\cmidrule(lr){0-5}
Time & 3.15                                                          & 1.80                                                           & 2.32                                                         & 2.19                                                         & \textbf{0.392} \\
Speed-up & 1x                                                               & 1.75x                                                             & 1.36x                                                           & 1.44x                                                           & \textbf{8.04x}   \\ 
\bottomrule
\end{tabular}
}
\vspace{-5pt}
\end{table}

We further compare the execution speed of \modelname with two data-driven models, SGAN and SR-LSTM. 
\tabref{tab:speed-comparison} shows that \modelname can run much faster than these models, specifically, 8.04x faster than SGAN. 
This efficiency enables \modelname to support large-scale applications such as traffic crowd simulation and sophisticated planning, to be demonstrated next. 

\section{Down-stream Applications}
Now we apply \modelname to  two down-stream tasks. In traffic simulation, we create a \modelname instance for each traffic agent and directly use the predicted motion for simulation. 
In real-time planning for autonomous driving, we use \modelname as the planner's sub-module to predict the motions of interfering agents.

\subsection{\modelname for Urban Traffic Simulation}

\begin{figure}[!t]
\centering                                                              
\begin{tabular}{ccc}
\hspace{-10pt}
\Fbox{\includegraphics[width=0.16\textwidth]{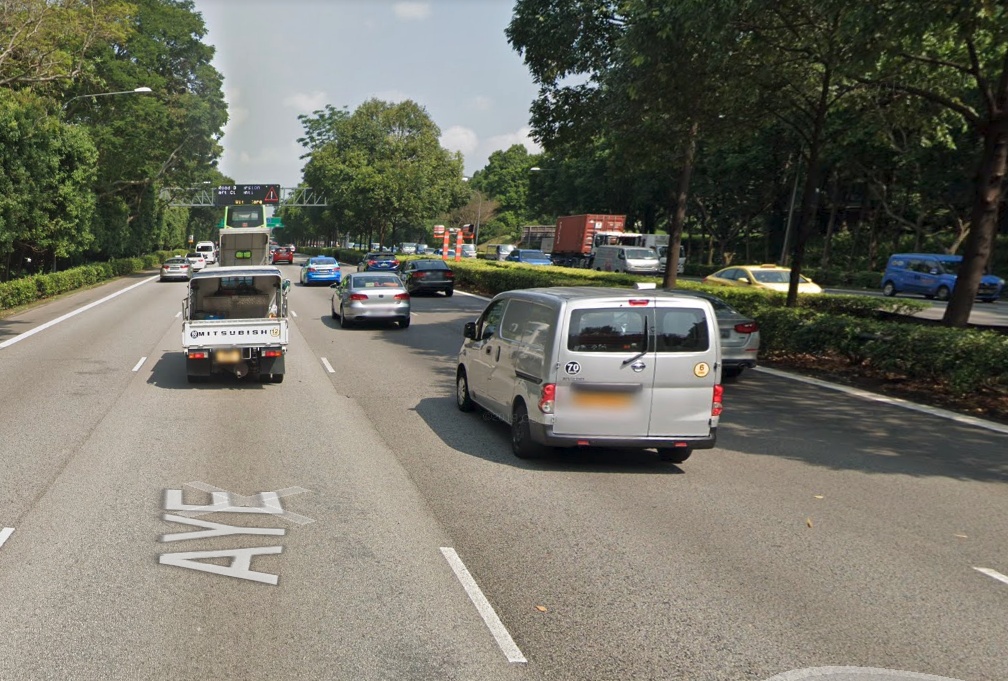}} & 
\hspace{-13pt}
\Fbox{\includegraphics[width=0.16\textwidth]{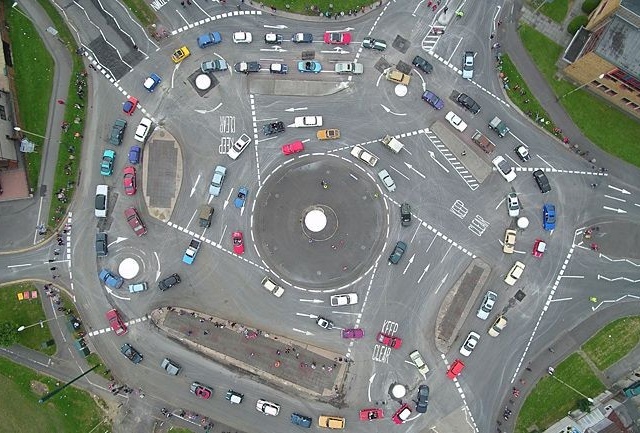}} & 
\hspace{-13pt}
\Fbox{\includegraphics[width=0.16\textwidth]{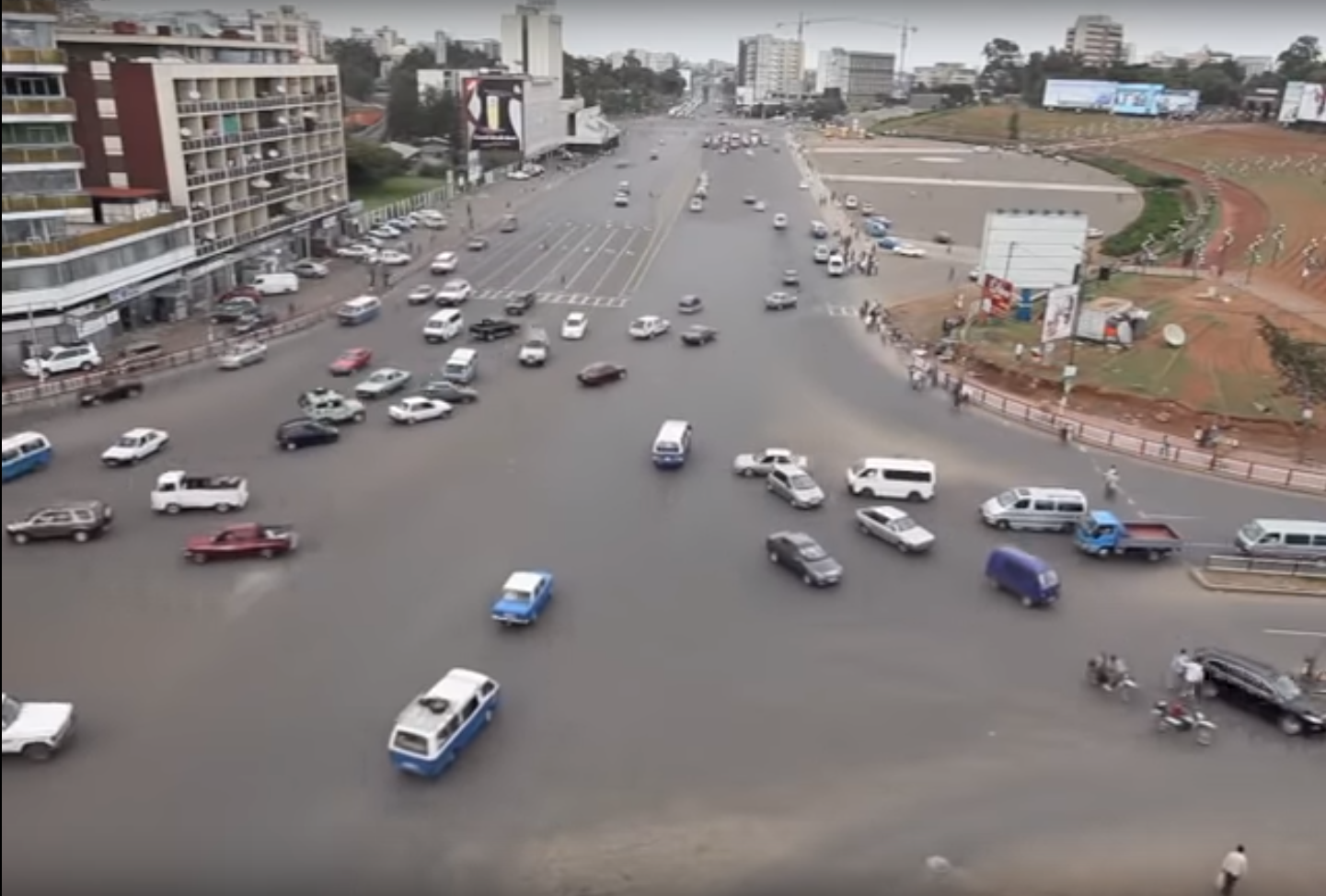}}  \\
\hspace{-10pt}
\Fbox{\includegraphics[width=0.16\textwidth]{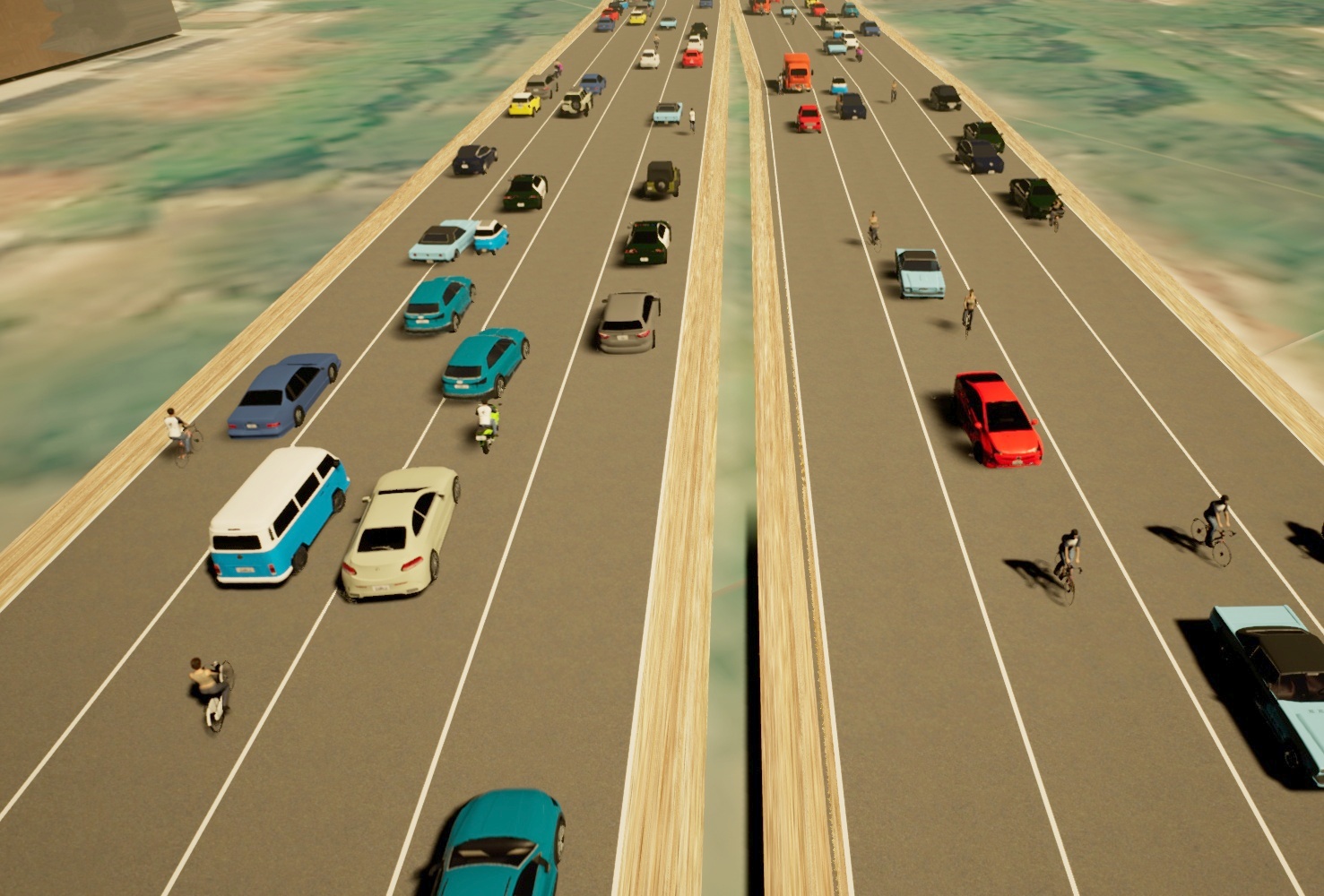}} & 
\hspace{-13pt}
\Fbox{\includegraphics[width=0.16\textwidth]{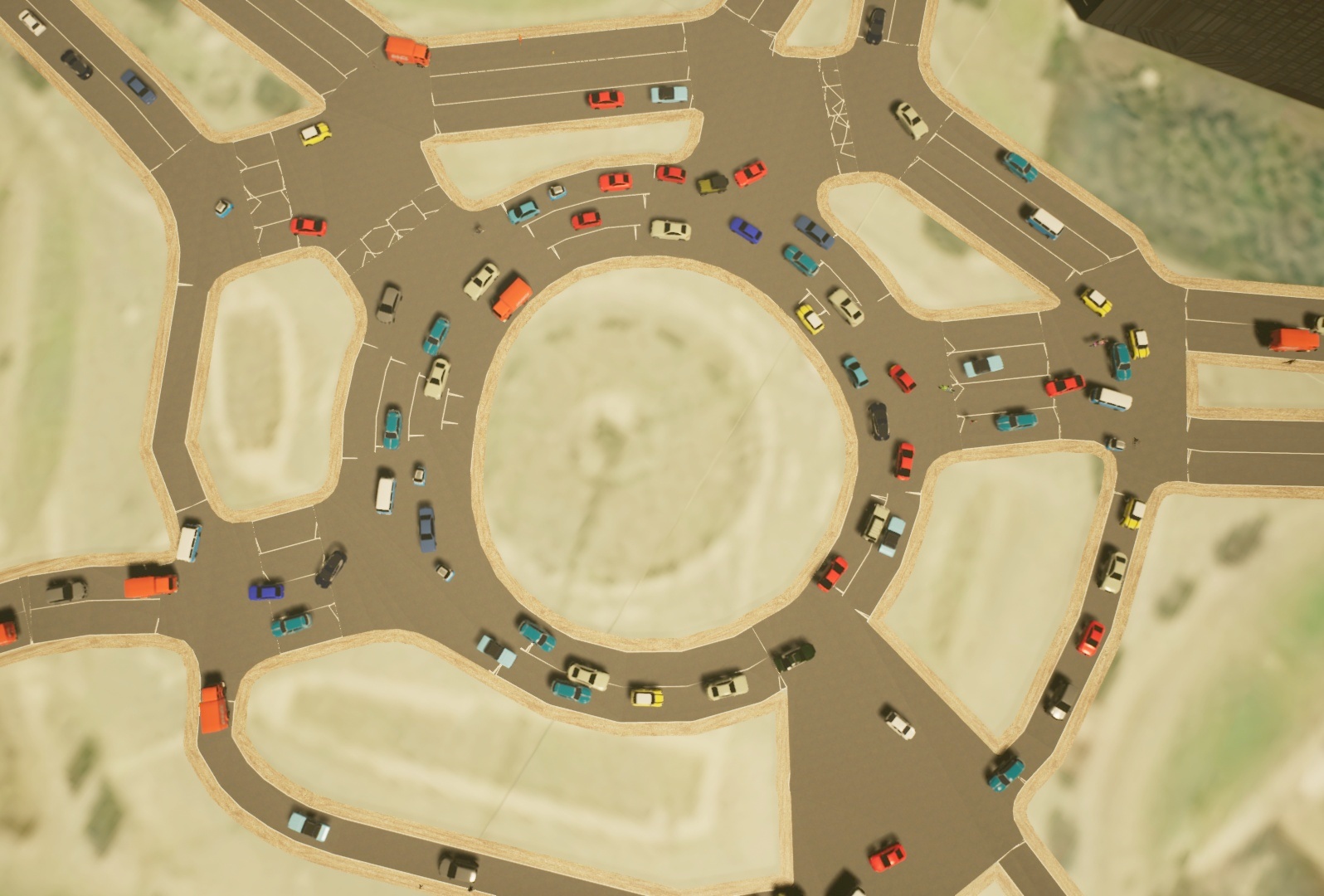}} & 
\hspace{-13pt}
\Fbox{\includegraphics[width=0.16\textwidth]{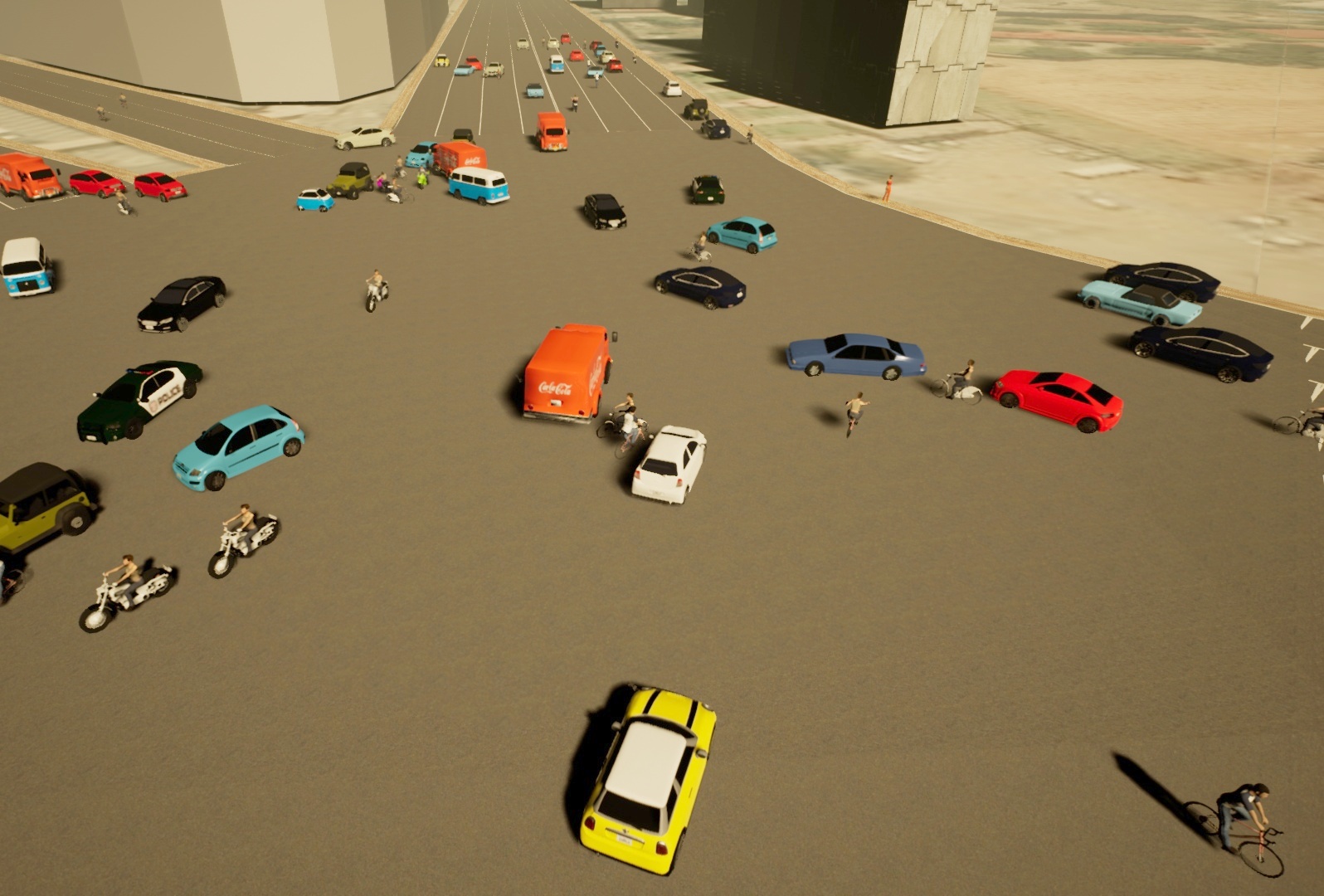}}  \\
\hspace{-10pt}{\footnotesize Highway}  & \hspace{-10pt}{\footnotesize Magic Roundabout}  & \hspace{-10pt}{\footnotesize Meskel Square}\\  
\hspace{-10pt}{\footnotesize Singapore}  & \hspace{-10pt}{\footnotesize Swindon, UK}  & \hspace{-10pt}{\footnotesize Addis Ababa, Ethiopia} 
\end{tabular}  
\captionof{figure}{GAMMA supports the SUMMIT simulator to generate urban traffic at any  worldwide location  available in the OpenStreetMap.}        
\label{fig:benchmarks}     
\vspace{-10pt}
\end{figure}

\rvs{
This section demonstrates the integration of \modelname with an open-source simulator, \simname \cite{cai2020summit}, to simulate massive mixed traffic on urban roads. 
\simname constructs virtual on-road driving scenes at any real-world location supported by the OpenStreetMap database \cite{OSM}. For each traffic agent in the scene, we use \modelname to control its motion conditioned on the road context. The entire scene contains $\approx100$ \modelname-controlled agents, each initialized with a random location, a random type, and unique \internalstates. Behaviors of all agents are updated in real-time at a rate of at least $20 HZ$.

The flexibility and efficiency of \modelname enable simulating realistic, dense traffic scenes in real-time. Agents in a crowd conduct their intentions smoothly and avoid collision collaboratively; All motions conform with the geometrical and kinematics constraints of the respective agents and the road structures; Individuals have versatile driving styles produced by sampling different intentions, responsibility levels, and attention levels. \figref{fig:benchmarks} shows sample simulations for three locations in Asia, Europe, and Africa. Simulation clips are included in the accompanying video (\href{https://youtu.be/teZJWlh8ZqI}{link}).
}

\subsection{\modelname for Real-Time Planning in Autonomous Driving} \label{sec::planner}
\rvs{\modelname can further support real-time planning under uncertainty for driving among challenging urban scenes. In this section, we consider driving in dense and unregulated urban traffic, where many agents interact with each other with no proper regulation by the traffic rules. A robot vehicle is deployed in the scenes to reach a sampled goal location.
To drive successfully through the crowd, the robot has to: 1) understand intentions and interactions of other agents to predict their motions accurately; 2) anticipate future risks of collision according to the predictions and plan the robot actions to avoid them ahead of time; 3) conditioned on the above, optimize the efficiency and smoothness of driving.

We plan the driving path of the vehicle using Hybrid A*, and formulate the acceleration control problem\footnote{Acceleration at each step is chosen from $\{\SI{3}{\meter \per \second^2}, 0, \SI{-3}{\meter \per \second^2}\}$, respectively. The maximum speed of the robot vehicle is $\SI{6}{\Ms}$.} as  a  partially  observable  Markov  decision process (POMDP) \cite{Kaelbling_1998}.
We argue that the main hazards in an unregulated traffic crowd come from the uncertainty over driver intentions and complex interactions. We tackle this challenge using belief-space planning with \modelname as the forward simulation model. 

Our POMDP planner treats the intentions of nearby agents as hidden state variables and tracks them using beliefs, i.e., probability distributions over states. The planner uses \modelname to predict how each exo-agent behaves in future situations when taking a particular intention. Starting from the current belief for all nearby agents, the planner performs Monte Carlo (MC) simulations to predict future outcomes, \ie, state transitions of the scene and noisy observations of them. Each MC simulation uses a specific sequence of robot accelerations and a sampled set of intentions of other agents. Conditioned on them, it predicts the future motions of all agents.

A tree of sampled future outcomes is yielded by repeating MC simulations using many randomly sampled scenarios and simulating all possible acceleration sequences of the robot under them. It is referred to as the ``belief tree''. Each node of the tree represents a future state of the scene and a future belief over agents' intentions. Each belief is assigned a reward according to the collision risk, driving progress, and driving smoothness assessed at the node.
To output the optimal action at the current belief, the planner applies Bellman’s principle across the entire belief tree to estimate the value of future beliefs and all action candidates to be taken there. Then, the optimal action for the current step with the best long-term outcome is extracted at the root of the tree. The process is commonly known as belief tree search, a standard approach to solving POMDPs online \cite{Kaelbling_1998}. We have particularly used a variant of DESPOT \cite{luo2019importance} to build our acceleration controller. 
The algorithm executes in an any-time fashion. It re-plans for every time step at a rate of 3~Hz, interleaving belief update, planning, and action execution. At each step, the planner searches hundreds of nodes to make reliable decisions, requesting thousands of predictions.
}

\begin{figure}[!t]
\centering                                                              
% \hspace{-5pt}
\includegraphics[width=0.165\textwidth]{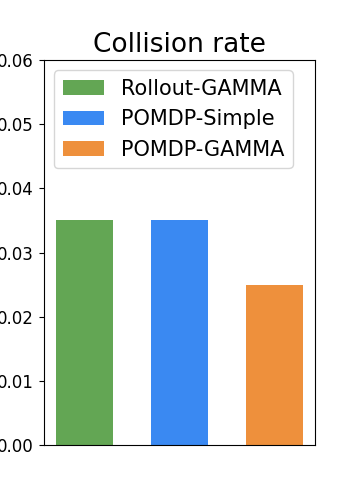}
\hspace{-13pt}
\Fbox{\includegraphics[width=0.165\textwidth]{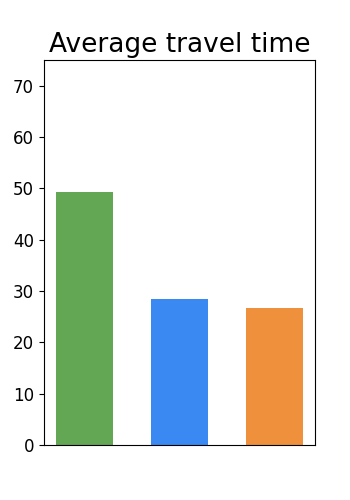}}
\hspace{-13pt}
\Fbox{\includegraphics[width=0.165\textwidth]{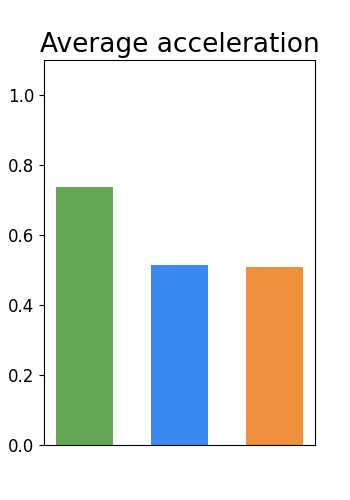}}

\vspace{-10pt}
 \captionof{figure}{Performance comparison of planning algorithms in the SUMMIT simulator for autonomous driving. 
}        
\label{fig:driving}     
\vspace{-10pt}
\end{figure}

We test the driving performance of the planner under the three sample scenes in \simname.
\figref{fig:driving} compares the performance of our planner (POMDP-GAMMA) with a simple roll-out planner using the same \modelname prediction model (Rollout-GAMMA) and a POMDP planner using a simpler prediction model assuming other vehicles follow their intended lanes with constant speeds (POMDP-Simple). The results show that both POMDP planning and accurate motion modeling are important for driving safety, efficiency, and smoothness. POMDP planning with \modelname achieves the lowest collision rate with the least time to reach goal locations, undertaking the minimum accelerations. Sample runs are included in the accompanying video (\href{https://youtu.be/teZJWlh8ZqI}{link}).

\section{conclusion}
\rvs{This paper presents \modelname, a fast, accurate, and flexible motion prediction model for heterogeneous traffic. 
Experimental results show that \modelname 
achieves strong performance in both prediction accuracy and computational efficiency on a set of standard benchmark datasets. 
These advantages  enable the successful application of \modelname to large-scale urban traffic simulation and real-time planning under uncertainty for autonomous driving.}

Our future work will focus on the following aspects. Currently, \modelname relies on the inference module to recognize the intentions of agents from observed histories and then uses the optimization module to predict motion.  It would be interesting to investigate joint inference and optimization for improved accuracy. 
\rvs{We also plan to evaluate \modelname on more versatile driving datasets, particularly those providing access to the road context, so that \modelname can predict typical driving maneuvers, including vehicle following, overtaking, lane merging, \etc.}

\end{document}